\documentclass[]{spie}  

\usepackage{amsmath,amsfonts,amssymb}
\usepackage{graphicx}
\usepackage[multi-part-units=single]{siunitx}
\usepackage[colorlinks=true, allcolors=blue]{hyperref}
\usepackage{todonotes}
\usepackage{bm}

\title{Bioresorbable Scaffold Visualization in IVOCT Images Using CNNs and Weakly Supervised Localization}

\author[a]{Nils Gessert}
\author[a]{Sarah Latus}
\author[b]{Youssef S. Abdelwahed}
\author[b]{David M. Leistner}
\author[c]{Matthias Lutz}
\author[a]{Alexander Schlaefer}
\affil[a]{Institute of Medical Technology, Hamburg University of Technology, Am Schwarzenberg-Campus 3, 21073 Hamburg, Germany}
\affil[b]{Charit\'{e} – Universit\"atsmedizin Berlin, Hindenburgdamm 30, 12203 Berlin, Germany}
\affil[c]{Universit\"atsklinikum Schleswig-Holstein, Arnold-Heller-Straße 3, 24105 Kiel, Germany}

\authorinfo{Further author information: (Send correspondence to Nils Gessert)\\Nils Gessert: nils.gessert@tuhh.de}

\begin{document} 
\maketitle

\begin{abstract}
Bioresorbable scaffolds have become a popular choice for treatment of coronary heart disease, replacing traditional metal stents. Often, intravascular optical coherence tomography is used to assess potential malapposition after implantation and for follow-up examinations later on. Typically, the scaffold is manually reviewed by an expert, analyzing each of the hundreds of image slices. As this is time consuming, automatic stent detection and visualization approaches have been proposed, mostly for metal stent detection based on classic image processing. As bioresorbable scaffolds are harder to detect, recent approaches have used feature extraction and machine learning methods for automatic detection. However, these methods require detailed, pixel-level labels in each image slice and extensive feature engineering for the particular stent type which might limit the approaches' generalization capabilities. Therefore, we propose a deep learning-based method for bioresorbable scaffold visualization using only image-level labels. A convolutional neural network is trained to predict whether an image slice contains a metal stent, a bioresorbable scaffold, or no device. Then, we derive local stent strut information by employing weakly supervised localization using saliency maps with guided backpropagation. As saliency maps are generally diffuse and noisy, we propose a novel patch-based method with image shifting which allows for high resolution stent visualization. Our convolutional neural network model achieves a classification accuracy of $\SI{99.0}{\percent}$ for image-level stent classification which can be used for both high quality in-slice stent visualization and 3D rendering of the stent structure.
 
\end{abstract}


\section{INTRODUCTION}
\label{sec:intro}  

Coronary heart disease is one of the most frequent causes of death despite being treatable. To treat the obstructive plaques, stenting is commonly used with mostly metallic stents being used in the past. As metal stents come with the risk of late stent thrombosis and in-stent restenosis\cite{cutlip2001stent}, bioresorbable scaffolds such as bioresorbable vascular scaffolds (BVS) have gained popularity recently. After implantation and in later follow-up examinations, the stents have to be assessed by the medical expert in order to detect malapposition or assess endothelialisation. Typically, intravascular optical coherence tomography (IVOCT) is used as an imaging modality for stent analysis\cite{tearney2012consensus} as it provides high resolution images of the lumen and vessel walls. As a single IVOCT pullback contains hundreds of image slices to be assessed, manual evaluation is labor-intensive and time consuming. Therefore, automatic stent detection and visualization methods have been proposed, mostly for metallic stents\cite{wang2013automatic,ughi2012automatic,gurmeric2009new}. These methods largely rely on classic image processing to detect the high-intensity metal stent struts. For bioresorbable scaffolds, a classic approach has also been proposed \cite{wang2014automatic}. However, since these struts are less pronounced in IVOCT images and different types of scaffolds show different characteristics, recent approaches have used machine learning methods combined with handcrafted feature extraction for detection and visualization \cite{cao2018automatic}. While showing promising results, these methods require pixel-level image annotations to learn local detection of stent struts within the slices. This is, once again, time consuming and limits the potential dataset size and variability. Moreover, features need to be engineered for a specific stent type which might not be suitable for future stent variations which has already become evident during the transition from metal stents to bioresorbable scaffolds\cite{wang2014automatic,cao2018automatic}.

For this reason, we propose a novel deep learning-based method for stent visualization and potential detection using only image-level label annotations. A convolutional neural network (CNN) is trained to classify an IVOCT slice into the categories "metal stent", "bioresorbable scaffold" and "no device". This way of image-level labeling has been successful for IVOCT-based deep learning\cite{gessert2018} as it is fast and thus allows for larger datasets. Moreover, it is easily extensible to new stent types as a new class simply needs to be added to the learning problem and no new feature engineering is required. After training the model, we employ the concept of weakly supervised localization\cite{oquab2015object} to derive local stent strut information from the model. In particular, we compute saliency maps with guided backpropagation\cite{Springenberg.2014} which can be interpreted as a gradient image which shows the regions that were most important for the model's prediction. In our case, the trained network should have learned to focus on stent struts. However, saliency maps are generally diffuse and high quality localization from global information only is very challenging\cite{oquab2015object}. For this reason, methods such as SmoothGrad\cite{smilkov2017smoothgrad} have been proposed which are targeted at improved saliency map quality. As we found this method  to be insufficient for the problem at hand, we propose a new patch-based approach with image shifting which leads to high resolution, high quality saliency maps that can serve as a visualization. The approach is used for regularization during model training and also for generation of stitched and averaged, smooth saliency maps. 

In this paper, we introduce our method for BVS visualization. We show that it is effective when visualizing stents for assessment after apposition with struts at the tissue surface as well as for follow-up review of, e.g. endothelialisation, where struts are starting to decay within the vessel tissue. Moreover, we show visualization of classic metal stents and the very recent Fantom Encore bioresorbable scaffold. 

\section{METHODS AND MATERIALS}

\subsection{Dataset} \label{sec:data}

The dataset we use consists of $6300$ clinical 2D IVOCT images in polar representation acquired with a St. Jude Medical Ilumien OPTIS. The set contains $965$ images with a metal stent, $1992$ with a BVS, and $3371$ without any device. The slices were extracted from $62$ pullbacks. Each of the images is assigned the label "metal stent", "bioresorbable scaffold", or "no device". Note, that we include metal stents for the sake of completeness and the main focus is on the more challenging bioresorbable scaffolds which have gained a lot more relevance in todays clinical routines. We use approximately $\SI{70}{\percent}$ of the dataset for training and $\SI{30}{\percent}$ for testing. The images are separated by pullbacks, i.e., all images from a pullback are either entirely in the training set or the test set. In addition to this dataset, we consider several 2D image slices from the recent Fantom Encore bioresorbable scaffold for evaluation. We do not train on any images containing this stent and we use the image slices to show the generalization of our method. We use the IVOCT images in their polar representation, i.e., the image axes are the depth $d$ and the angle of acquisition $\theta$. For visualization, we transform the images to cartesian space using $x = d\cos(\theta)$ and $y = d\sin(\theta)$.

\subsection{Model Training}

We use a 2D CNN which takes the polar 2D IVOCT images as its input and outputs one of the three classes. The model is a Resnet50 that was pretrained on the ImageNet dataset\cite{He.2016}. We employ a preprocessing scheme that matches the evaluation strategy later on for saliency map generation. First, we cut off the last $200$ pixels along the depth dimension of the original polar images as they mostly contain noise. Then, we do not downsample the images any further, as typically done, but instead use small crops of size $224 \times 224$ or $160 \times 160$ out of the image of size $496 \times 776$. In this way, we maintain OCT's original high resolution. This method induces a regularizing effect as each crop will only contain few stent struts and never the entire structure. This ensures that the network does not fit to only a single strut which might be sufficient to identify the stent type. For further data augmentation, we randomly shift the polar image along the angle dimension which ensures that information at the image borders is not neglected or ignored by the model.  

As the dataset exhibits class imbalance, we weight the cross-entropy loss with the normalized inverse class frequency of each class, i.e., the loss for metal stent examples is weighted higher than the loss for normal images. We train the model using stochastic gradient descent with Adam\cite{Kingma.2014} with a starting learning rate of $l_r = 0.0001$, a batch size of $B = 40$ and a training time of $400$ epochs. We implement preprocessing, model training and saliency map generation using Tensorflow\cite{Abadi.2016}.

\subsection{Saliency Map Construction} \label{sec:saliency}

\begin{figure}
\centering
\includegraphics[width=1.0\textwidth]{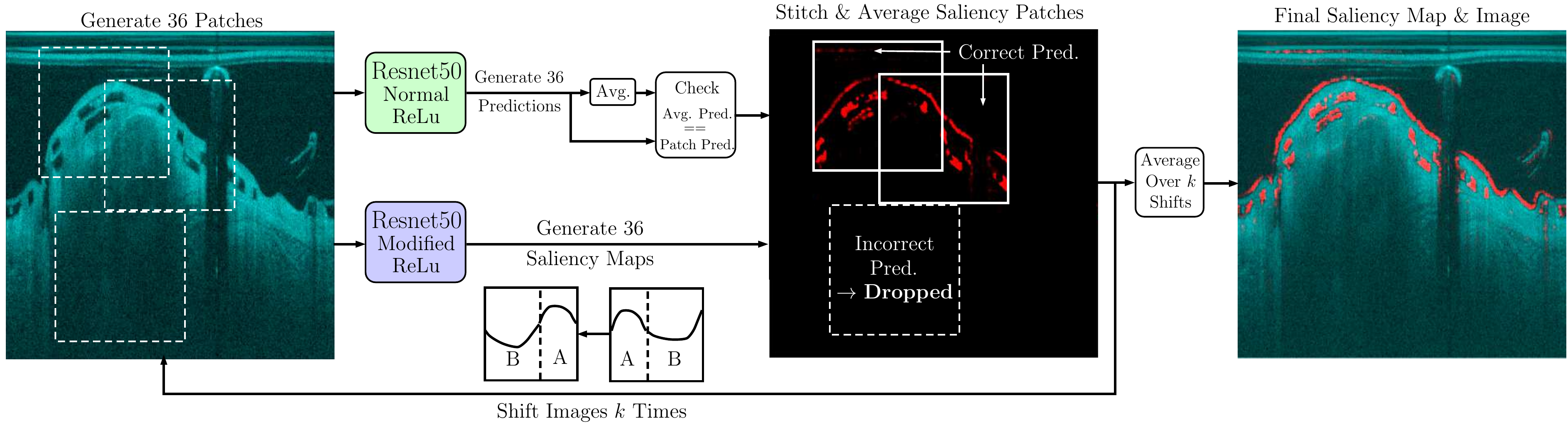}
\caption{The proposed evaluation strategy for saliency map generation. For training, patches are sampled similarly and shifting is used as a data augmentation technique. Note, that shifting in polar coordinate space resembles rotation in cartesian space. For details, see Section~\ref{sec:saliency}.}
\label{fig:eval}
\end{figure}

After training, we use the model to construct saliency maps for stent visualization. The generation process is depicted in Figure~\ref{fig:eval}. First, $36$ overlapping patches are cropped from the original image. Then, with the normal network, predictions are obtained for each patch. These predictions are averaged for a global image prediction with respect to the stent type. In addition, the $36$ patches are used for saliency map generation using guided backpropagation\cite{Springenberg.2014} which requires a modification of the ReLu backwards pass. Then, the saliency map patches are assembled into an entire image once again. During this process, only patches are considered that match the global, average prediction for the current image. This avoids noise being added into the image in unwanted locations. Moreover, we cut off $\SI{10}{\percent}$ of the saliency patches' borders as we observed noise and artifacts in those regions. Afterwards, the process is repeated $k$ times with shifted versions of the original polar image along the angle dimension. This shift represents a rotation of the image in cartesian space. The $k$ saliency maps are then averaged into a final saliency map. This process smooths the maps and makes sure that stent struts at the image borders are captured as well. As convolutions are invariant towards translation, we found $k=3$ to be sufficient to take care of border effects. For comparison, we also consider a naive approach where the model is trained on whole, downsampled images and the saliency maps are generated by a single pass over the entire image.
For visualization, we use the negative saliency map, unless indicated otherwise. I.e., we only plot values from the saliency map that have a negative value as we found this variant to highlight the stent struts effectively.

\section{RESULTS}

\begin{table}
\centering
\begin{tabular}{l l l l}
	& Full Images & Patch Size $224\times 224$ & \textbf{Patch Size} $\bm{160\times 160}$  \\ \hline 
    Accuracy & $0.979$ & $0.986$ & $\bm{0.990}$ \\
    AUC & $0.997$ & $0.998$ & $\bm{0.999}$ \\         
    F1-Score & $0.979$ & $0.986$ & $\bm{0.989}$  \\      
\end{tabular}
\caption{Stent classification results for three different approaches. Accuracy and area-under-curve (AUC) represent the mean value over all classes.}
\label{tab:res}
\end{table}

First, we report classification accuracy results for several model variations in Table~\ref{tab:res}. We consider two variants of our approach with patches of size $224\times 224$ and $160\times 160$. Moreover, we provide results for the whole image approach where the entire image is downsampled instead of using high resolution patches. There is only a small difference between the models with the patch-based approaches performing slightly better. Overall, the stent classification accuracy is very high.

\begin{figure}
\centering
\includegraphics[width=1.0\textwidth]{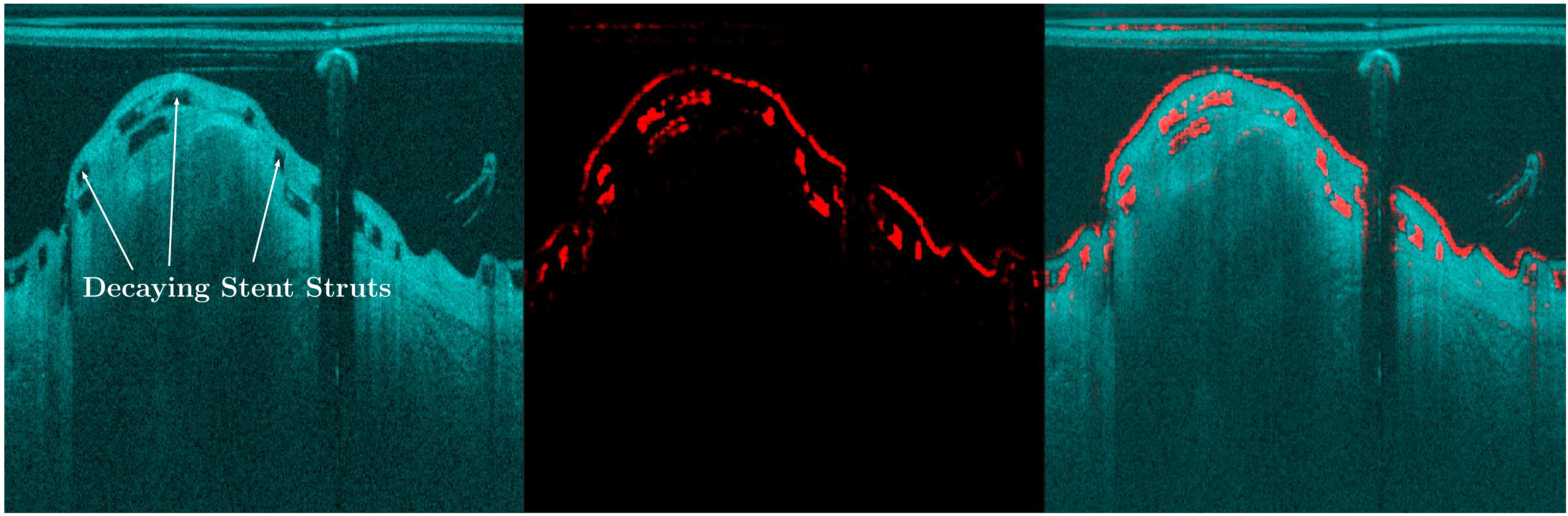}
\includegraphics[width=0.9935\textwidth]{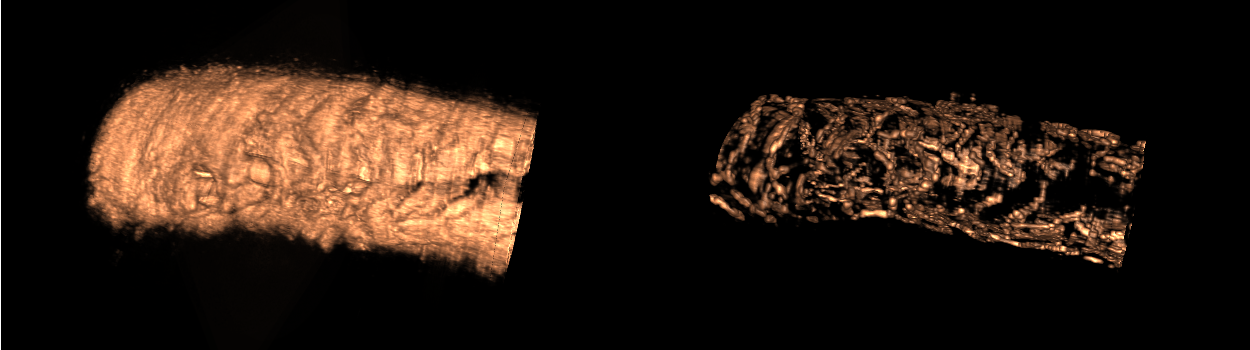}
\caption{Example for our saliency map-based visualization for decaying stent struts. Top, the visualization within the slice is shown. Left, the original IVOCT image in polar representation is shown. In the center, the generated saliency map is shown. Right, the overlaid image is shown. The saliency map is shown in red. The original IVOCT image is shown in cyan. Bottom, a 3D rendering of our visualization technique is shown. Left, the normal rendered pullback is shown. Right, the stacked, rendered saliency maps are shown.}
\label{fig:pol_sal1}
\end{figure}

Next, Figure~\ref{fig:pol_sal1} shows the visualization of an IVOCT image, its saliency map and the overlaid image for the case of struts being covered by tissue in a follow-up examination. The saliency map clearly delinates both the lumen and the stent struts. Furthermore, the figure shows a 3D visualization of a part of a pullback. Here, the IVOCT images were transformed to cartesian space and stacked next to each other. Our visualization clearly shows the stent's grid structure across slices.

\begin{figure}
\centering
\includegraphics[width=1.0\textwidth]{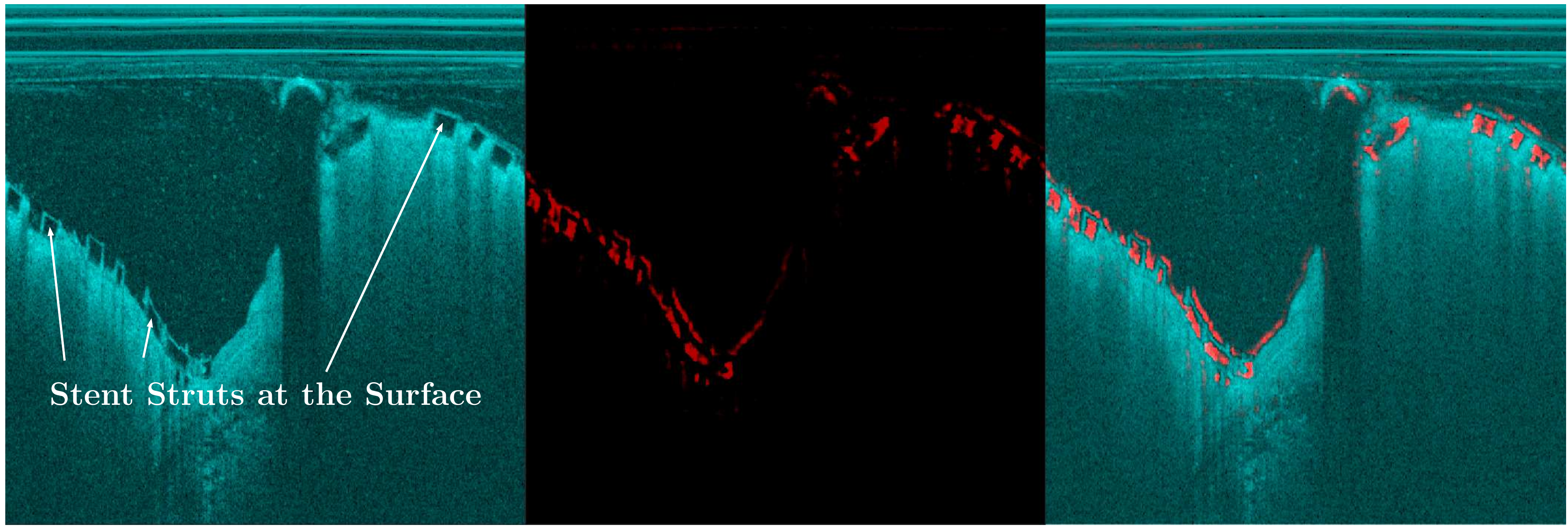}
\includegraphics[width=0.995\textwidth]{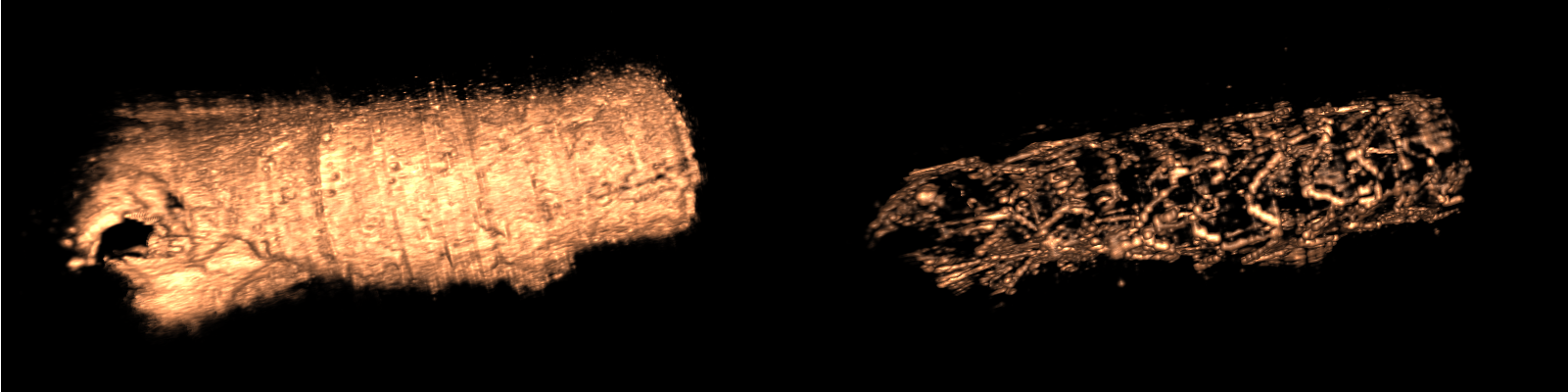}
\caption{Example for our saliency map-based visualization for stent struts at the surface. Top, the visualization within the slice is shown. Left, the original IVOCT image in polar representation is shown. In the center, the generated saliency map is shown. Right, the overlaid image is shown. The saliency map is shown in red. The original IVOCT image is shown in cyan. Bottom, a 3D rendering of our visualization technique is shown. Left, the normal rendered pullback is shown. Right, the stacked, rendered saliency maps are shown.}
\label{fig:pol_sal2}
\end{figure}

In addition, Figure~\ref{fig:pol_sal2} shows the same visualization for the case of stent struts at the tissue surface, which is commonly the case directly after implantation. Again, the saliency maps provide a meaningful visualization both within the slice and also in a 3D rendering.

\begin{figure}
\centering
\includegraphics[width=1.0\textwidth]{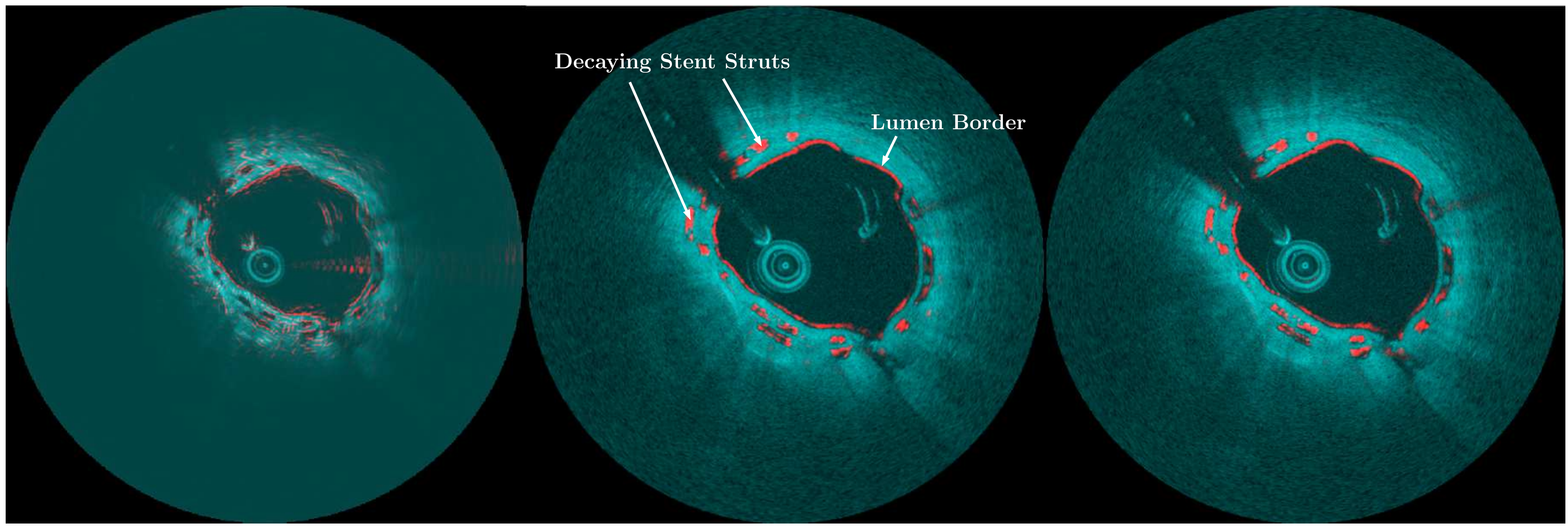}
\caption{Comparison of saliency maps between the full image-based approach and the cropped strategy. Left, the full image-based approach is shown. In the center, the patch-based approach with $224\times 224$ is shown. Right, the patch-based approach with $160\times 160$ is shown. The images were transformed to cartesian space for visualization. The saliency map is colored in red. The original IVOCT image is colored in cyan.}
\label{fig:comp}
\end{figure}

Furthermore, we present several visualizations from variations of our approach. Figure~\ref{fig:comp} shows the resulting saliency maps for the three different training scenarios. Clearly, our approach substantially improves saliency map quality with clear outlines of the stent struts and also the lumen border. 

\begin{figure}
\centering
\includegraphics[width=1.0\textwidth]{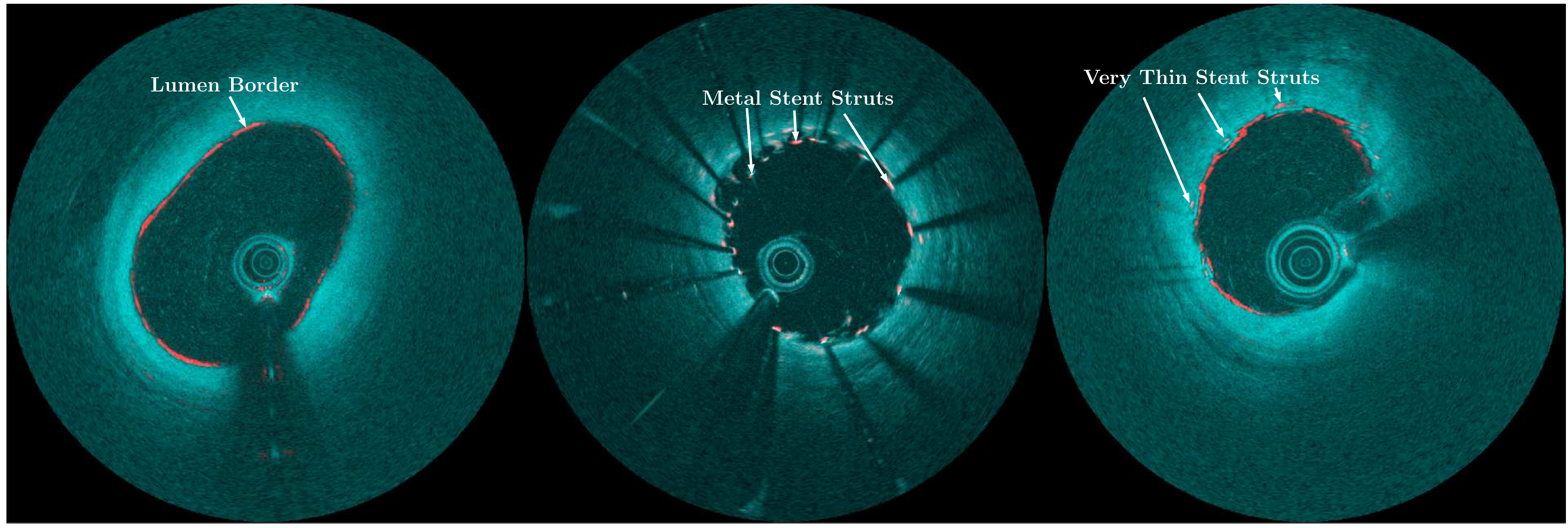}
\caption{Comparison of saliency maps between no stent (left), a metal stent (center) and the Fantom Encore bioresorbable scaffold (right). Note, that for the metal stent visualization we used maximum saliency maps instead of minimum.  The images were transformed to cartesian space for visualization. The saliency map is colored in red. The original IVOCT image is colored in cyan.}
\label{fig:comp_stent}
\end{figure}

Last, we show a comparison between visualization without any stent, a metal stent and the new Fantom Encore bioresorbable scaffold in Figure~\ref{fig:comp_stent}. The figure shows that our method also works well for metal stent and even generalizes to a new stent model whose struts are signifcantly thinner than the BVS model in the training set. Also, it is notable that our method achieves a smooth lumen segmentation when no stent is present in the image.

\section{DISCUSSION AND CONCLUSION}

In this paper we present a novel deep learning-based method for bioresorbable scaffold visualization. The method does not require any pixel-level annotations and local image information is derived from a global image label only. In particular, we train a convolutional neural network that predicts the type of stent that is generally visible in the IVOCT image. Then, we follow the concept of weakly supervised localization\cite{Oquab.2014} and derive fine-grained image information from saliency maps with guided backpropagation\cite{Springenberg.2014}. As saliency maps are generally diffuse, we propose a patch-based method with image shifting for smoothed saliency maps. The concept is also useful for training where it acts as a regularization technique which slightly improves stent classification accuracy, see Table~\ref{tab:res}. In general, the classification performance is very high which indicates that the learning problem is well defined and sufficiently solved with our model. The resulting visualizations in Figure~\ref{fig:pol_sal1} and Figure~\ref{fig:pol_sal2} show that our method is able to capture bioresorbable stent struts both for detection at the surface and in later stages of decay. The former is clinically relevant as malapposition needs to be detected immediately after implantation with struts at the surface. The latter is important for follow-up examination where, e.g. endothelialisation needs to be confirmed. Also, it is notable that the 3D visualization of our saliency maps accurately shows the expected stent grid structure which shows that the visualization is consistent across slices. In addition, we showed that our new method significantly improves saliency map quality compared to a standard saliency map approach with full images as the network input. Last, we found that our method also works well for metal stents and notably also the new Fantom Encore bioresorbable scaffold. The visualization also works for the latter despite the stent struts being very small. Also, we did not include any stent images of this type during training which shows that our method is also able generalize to new, similar stent types. Even if an entirely new stent type was added to our method, adjustment would be very easy as only an additional class needs to be added to the learning problem. No manual feature engineering is required. 
For future work, our successful visualization method can be extended to quantitative analysis, e.g. for automatic stent malapposition detection. This extension should be feasible as the model also appears to have learned a reasonable lumen segmentation, see Figure~\ref{fig:comp_stent}. Adding an explicit differentiation between stents and lumen border could allow for measurement of the distance between struts and the lumen. Furthermore, our method could be extended with a larger dataset and more stent types.




\bibliography{egbib.bib} 
\bibliographystyle{spiebib} 

\end{document}